%% file: main.tex
\documentclass[11pt]{article}
\usepackage{fullpage}
\makeatletter
\long\def\@makecaption#1#2{
  \vskip 0.8ex
  \setbox\@tempboxa\hbox{\small {\bf #1:} #2}
  \parindent 1.5em 
  \dimen0=\hsize
  \advance\dimen0 by -3em
  \ifdim \wd\@tempboxa >\dimen0
  \hbox to \hsize{
    \parindent 0em
    \hfil 
    \parbox{\dimen0}{\def\baselinestretch{0.96}\small
      {\bf #1.} #2
     
    } 
    \hfil}
  \else \hbox to \hsize{\hfil \box\@tempboxa \hfil}
  \fi
}
\makeatother

\usepackage{times,enumitem,hyperref,booktabs,multicol,blindtext,url,parskip}

\usepackage{pifont}

\usepackage{adjustbox}
\usepackage{amsfonts}
\usepackage{amsmath}
\usepackage{amsthm}
\usepackage{amssymb}
\usepackage{microtype}
\usepackage{graphicx}
\usepackage{natbib}
\usepackage{fancyhdr}
\usepackage{color}
\usepackage{algorithm}
\usepackage{forloop}
\input{macros.tex}

\usepackage{multirow}
\usepackage{enumerate}
\usepackage{algpseudocode}
\usepackage{rotating}
\usepackage{makecell}
\usepackage{bm}
\usepackage{wrapfig}

\usepackage{graphicx}
\usepackage{subfigure}
\usepackage{multicol}
\usepackage{pdfpages}
\usepackage{xspace}
\usepackage{float}
\usepackage{threeparttable}

\def\ie{\textit{i.e.}}
\def\eg{\textit{e.g.}}
\def\etal{\textit{et~al.}}

\usepackage[capitalize]{cleveref}
\crefname{section}{Sec.}{Secs.}
\Crefname{section}{Section}{Sections}
\Crefname{table}{Table}{Tables}
\crefname{table}{Tab.}{Tabs.}

\def\shownotes{0}  
\ifnum\shownotes=1
\newcommand{\authnote}[2]{[#1: #2]}
\else
\newcommand{\authnote}[2]{}
\fi

\definecolor{mypink}{RGB}{219, 48, 122}

\hypersetup{
  colorlinks,
  citecolor=violet,
  linkcolor=red,
  urlcolor=blue}

\begin{document}

\abovedisplayskip=8pt plus0pt minus3pt
\belowdisplayskip=8pt plus0pt minus3pt

\begin{center}
  \vspace*{-1cm}
  {\LARGE Mettle: Meta-Token Learning for Memory-Efficient \\ Audio-Visual Adaptation
  \vspace{0.12cm}
  ~} \\
  \vspace{.4cm}
  {\large Jinxing Zhou$^{1}$,
        Zhihui Li$^{3,*}$, 
        Yongqiang Yu$^{1}$,
        Yanghao Zhou$^{4}$,
        Ruohao Guo$^{5}$, 
        Guangyao Li$^{6}$, \\
        Yuxin Mao$^{7}$,
        Mingfei Han$^{1}$,
        Xiaojun Chang$^{1,3}$,
        Meng Wang$^{2,*}$
  } \\

  \vspace{.2cm}
$^{1}$MBZUAI \qquad
$^{2}$Hefei University of Technology  \qquad
$^{3}$University of Science and Technology of China \\
$^{4}$National University of Singapore \qquad
$^{5}$Peking University \qquad
$^{6}$Tsinghua University \qquad
$^{7}$OpenNLP Lab

\renewcommand{\thefootnote}{}
\footnotetext{* Corresponding authors (lizhihuics@ustc.edu.cn, eric.mengwang@gmail.com)} 
\renewcommand{\thefootnote}{\arabic{footnote}}

\end{center}

\begin{abstract}
Mainstream research in audio-visual learning has focused on designing task-specific expert models, primarily implemented through sophisticated multimodal fusion approaches.
Recently, a few efforts have aimed to develop more task-independent or universal audiovisual embedding networks, encoding advanced representations for use in various audiovisual downstream tasks.
This is typically achieved by fine-tuning large pretrained transformers, such as Swin-V2-L and HTS-AT, in a parameter-efficient manner through techniques such as tuning only a few \textit{adapter} layers inserted into the pretrained transformer backbone.
Although these methods are parameter-efficient, they suffer from significant training memory consumption due to gradient backpropagation through the deep transformer backbones, which limits accessibility for researchers with constrained computational resources.
In this paper, we present \textbf{Met}a-\textbf{T}oken \textbf{Le}arning (Mettle), a simple and memory-efficient method for adapting large-scale pretrained transformer models to downstream audio-visual tasks.
Instead of sequentially modifying the output feature distribution of the transformer backbone, Mettle utilizes a lightweight \textit{Layer-Centric Distillation (LCD)} module to distill in parallel the intact audio or visual features embedded by each transformer layer into compact meta-tokens.
This distillation process considers both pretrained knowledge preservation and task-specific adaptation.
The obtained meta-tokens can be directly applied to classification tasks, such as audio-visual event localization and audio-visual video parsing.
To further support fine-grained segmentation tasks, such as audio-visual segmentation, we introduce a \textit{Meta-Token Injection (MTI)} module, which utilizes the audio and visual meta-tokens distilled from the top transformer layer to guide feature adaptation in earlier layers.
Extensive experiments on multiple audiovisual benchmarks demonstrate that our method significantly reduces memory usage and training time while maintaining parameter efficiency and competitive accuracy.

\end{abstract}


\section{Introduction}\label{sec:intro}

\begin{figure*}[t]
  \centering
\includegraphics[width=1\textwidth]{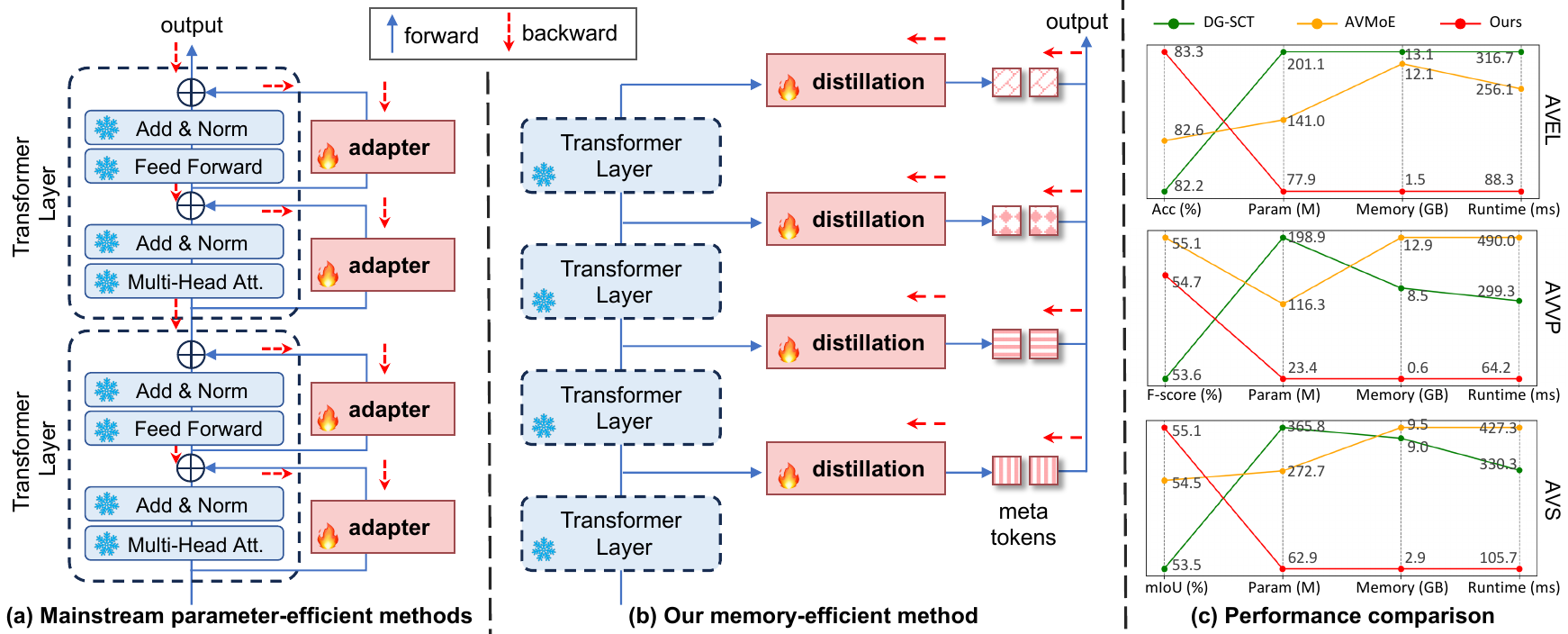}
   \caption{\textbf{(a)} Mainstream parameter-efficient methods for audiovisual adaptation insert learnable adapters within each frozen transformer layer. This alters the original feature output, inducing heavy memory overhead during backpropagation. \textbf{(b)} The core idea of our memory-efficient method aims to generate compact meta-tokens via parallel distillation from each transformer layer, preventing gradient propagation through the transformer backbone.
   \textbf{(c)} Comparison with prior state-of-the-art methods (\eg, DG-SCT~\cite{duan2024dgsct} and AVMoE~\cite{cheng2024avmoe}) in terms of accuracy, trainable parameters, memory and runtime during training on the AVEL~\cite{tian2018audio}, AVVP~\cite{tian2020unified}, and AVS~\cite{zhou2022avs} tasks. The multi-source setting of AVS is illustrated. 
   Swin-V2-L~\cite{liu2022swinv2} and HTS-AT~\cite{chen2022htsat} are used as the visual and audio transformer backbones, respectively.
   Memory and runtime values represent the per-sample consumption during model training (\ie, batch size = 1).
   }
   \label{fig:intro}
\end{figure*}

Audio-visual learning~\cite{wei2022learning} has received increasing attention in recent years, aiming to enable machines to mimic human intelligence in processing and analyzing complex audiovisual signals from real-world scenarios.
Prior studies primarily focus on developing specialized expert models for individual audiovisual learning tasks, such as classification~\cite{tian2018audio,tian2020unified} and segmentation~\cite{zhou2022avs} problems.
To enhance these tasks, various task-oriented modules have been proposed to improve the audio-visual \textit{late} fusion ~\cite{lin2019dual,zhou2021positive,xu2020cross,xia2022cross,yu2021mm,tian2020unified,mo2022multi,zhou2022avs,li2023catr}.

Recent advances~\cite{lin2023lavish,duan2024dgsct,cheng2024avmoe} in audio-visual learning aim to enhance more task-universal \textit{early} audiovisual feature encoding stage by fine-tuning large-scale pretrained transformer backbones (\eg, ViT-L-16~\cite{dosovitskiy2020vit} and Swin-V2-L~\cite{liu2022swinv2}) using a \textit{parameter-efficient} paradigm.
As shown in Figure~\ref{fig:intro}a, mainstream parameter-efficient methods, such as LAVisH~\cite{lin2023lavish}, DG-SCT~\cite{duan2024dgsct}, and AVMoE~\cite{cheng2024avmoe}, typically insert adapter~\cite{houlsby2019adapter} layers into the residual connection branch of each transformer layer.
These adapter layers are designed to learn task-related representations for downstream tasks.
Since the parameters of the pretrained transformer layers remain frozen while only the inserted adapter layers are trainable, these methods ensure parameter efficiency.
However, they suffer from \textit{memory inefficiency}. 
As illustrated in Figure~\ref{fig:intro}a, the output of each transformer layer is changed and gradients need to be propagated through the entire model to update adapter parameters.
This leads to substantial memory overhead and significantly increased optimization time during training.
For example, the prior state-of-the-art AVMoE~\cite{cheng2024avmoe} requires 12.1GB of GPU memory for training on the AVE~\cite{tian2018audio} dataset, even with a batch size of 1.
Such limitation prohibits the use of larger batch sizes for faster training and poses challenges for researchers with constrained computational resources.

In this paper, we explore \textit{\textbf{memory-efficient audio-visual adaptation}}, which aims to adapt large-scale pretrained transformer models for downstream audiovisual learning tasks in a memory-efficient manner, while maintaining parameter efficiency and competitive accuracy.
The memory inefficiency of prior models~\cite{lin2023lavish,duan2024dgsct,cheng2024avmoe} arises from the \textit{sequential} processing of intermediate transformer features from bottom to top layers.
We consider enforcing the model learning process in \textit{parallel} with transformer layers.
Since transformer backbones comprising deep layers are pretrained on large datasets, they can capture generalizable patterns and structural regularities in feature embedding~\cite{raghu2021vision,cordonnier2019relationship}.
Thus, the encoded intermediate features retain useful information, allowing them to be directly and layer-independently leveraged for downstream adaptation.

Motivated by this, we propose a
\textbf{Layer-Centric Distillation (LCD)} module, which aims to distill useful representations from each pretrained transformer layer into a set of \textbf{meta-tokens}, as illustrated in Figure~\ref{fig:intro}b.
We refer to them as \textit{meta}-tokens because they are highly compact, with significantly fewer elements than the source audio or visual tokens embedded by pretrained transformer layers.
Specifically, the meta-tokens are first initialized as learnable vectors and then updated through iterative distillation from pretrained modality-specific tokens.
LCD performs this distillation through two key mechanisms (see Figure~\ref{fig:framework}a).
First, the learnable meta-tokens interact with pretrained audio/visual tokens by modeling their cross-attentions~\cite{dosovitskiy2020vit}, querying {task-relevant} representations to refine the meta-tokens;
Meanwhile, a lightweight linear layer is directly applied to the original audio/visual tokens to better preserve effective {pretrained knowledge}.
Afterward, the obtained meta-tokens from different transformer layers are aggregated to comprehensively leverage diverse representations captured across layers.
This aggregation is implemented using parameter-free average pooling, yielding ultimate meta-tokens for downstream adaptation.
We apply LCD separately to the audio or visual modality to generate \textit{early} audio/visual meta-tokens. Existing sophisticated audio-visual interaction strategies can then be incorporated at \textit{late} stage to generate predictions for downstream tasks~\cite{duan2024dgsct,cheng2024avmoe}.

We dub this approach as \textit{\textbf{\underline{Met}a-\underline{T}oken \underline{Le}arning ({Mettle})}}, where the learned meta-tokens can be directly used for downstream classification tasks in audio-visual learning, such as audio-visual event localization (AVEL)~\cite{tian2018audio} and audio-visual video parsing (AVVP)~\cite{tian2020unified}.
Furthermore, we extend \textit{{Mettle}} to more complex segmentation tasks, such as audio-visual segmentation (AVS)~\cite{zhou2022avs}.
For such tasks, it is infeasible to generate pixel-level predictions solely from compact audio or visual meta-tokens.
We further propose the \textbf{Meta-Token Injection (MTI)} module, which reintegrates the distilled knowledge in meta-tokens into visual tokens (higher-resolution flattened features maps) encoded by pretrained transformers (illustrated in Figure~\ref{fig:framework}b).
Specifically, we apply the LCD module only to the final layer of the pretrained transformer backbone to obtain audio and visual meta-tokens containing high-level semantics.
These meta-tokens are then injected into the visual tokens from earlier layers through cross-modal and intra-modal dot-product attention~\cite{vaswani2017attention} modeling.
This process allows the hidden knowledge captured by audio or visual meta-tokens to enhance the representations of pretrained visual tokens, directing them toward task-related adaptations (\eg, highlighting audio-driven regions of interest in AVS task).

Our \textit{Mettle} approach avoids gradient backpropagation along the large transformer backbone, ensuring memory efficiency.
Moreover, the LCD module and the MTI module inserted into the transformer layer(s) are deliberately designed to be simple and lightweight, requiring only minimal learnable parameters. As a result, our model is both parameter- and memory-efficient.
We conduct extensive experiments on three representative audio-visual downstream tasks: AVEL~\cite{tian2018audio}, AVVP~\cite{tian2020unified}, and AVS~\cite{zhou2022avs}.
Figure~\ref{fig:intro}c provides a comprehensive comparison of our method with prior parameter-efficient works~\cite{duan2024dgsct,cheng2024avmoe}.
Our method achieves competitive or superior accuracy while significantly reducing the consumption of trainable parameters, training memory, and runtime.
For example, on the AVEL task, our method outperforms prior state-of-the-art AVMoE~\cite{cheng2024avmoe} by 0.7\% in accuracy, while using \textbf{45\%} fewer trainable parameters, \textbf{88\%} less memory, and \textbf{65\%} less training runtime.

To summarize, our main contributions are threefold:
\begin{itemize}
    \item We propose \textit{Mettle}, the first memory-efficient framework for adapting large-scale pretrained transformers to downstream audio-visual learning tasks.
    \item We introduce lightweight technical designs, including the Layer-Centric Distillation (LCD) and Meta-Token Injection (MTI), highlighting compact meta-token learning and utilization.
    \item We achieve superior performance across multiple audio-visual classification and segmentation benchmarks, maintaining competitive accuracy while ensuring high efficiency in parameters, memory usage, and training time.
\end{itemize}

\section{Related Work}\label{sec:related_work}

\subsection{Audio-Visual Learning}\label{sec:related_work_avl}
Audio-Visual Learning has made significant progress in recent years, with studies covering a range of tasks, such as classification~\cite{arandjelovic2018objects,cheng2020look,tian2018audio,tian2020unified,zhou2025dense,liu2025towards}, detection~\cite{zhao2018sound,chen2021localizing,hu2020discriminative,mahmud2024t}, segmentation~\cite{zhou2022avs,zhou2023avss,guo2024open,guo2025audio}, and text-involved captioning~\cite{tian2018attempt,shen2023fine,tian2019audio,mao2024tavgbench,kim2024avcap} or question answering~\cite{yun2021pano,li2022learning,li2024object,li2023progressive,li2024boosting,li2025patch} tasks.
In this work, we focus on three key audiovisual tasks.
The \textit{Audio-Visual Event Localization (AVEL)}~\cite{tian2018audio,lin2019dual,zhou2021positive,zhou2022contrastive,lin2020audiovisual,rao2022dual,xia2022cross,zhou2025towards} task aims to identify temporal segments where both audible and visible events occur, providing a fundamental understanding of audio-visual correspondence.
Since audio and visual signals are not always temporally or spatially aligned (\eg, the sounding objects may be out-of-view), researchers have further explored the \textit{Audio-Visual Video Parsing (AVVP)}~\cite{tian2020unified,wu2021exploring,cheng2022joint,gao2023collecting,lai2023modality,sardari2024coleaf,zhou2023improving,zhou2024label,zhou2024advancing,zhao2025multimodal} task.
Unlike AVEL, AVVP not only detects audio-visual events but also distinguishes events that are exclusively audible or visible.
Beyond these temporal classification tasks, the \textit{Audio-Visual Segmentation (AVS)}~\cite{zhou2022avs,li2023catr,gao2024avsegformer,liu2024audio,guo2024enhance,ma2024stepping,chen2024unraveling,zhoualoha} task focuses on the spatial comprehension of audiovisual scenes by generating pixel-level masks of sounding objects.
Traditional methods for these tasks primarily focus on the \textit{late} fusion of extracted audiovisual features.
In contrast, our work enhances \textit{early} feature encoding by adapting rich representations from large-scale pretrained transformers.

\subsection{Parameter-Efficient Transfer Learning}
Parameter-Efficient Transfer Learning (PETL) aims to adapt large pretrained foundation models to downstream tasks by fine-tuning only a small subset of parameters.
Initially developed in the natural language processing (NLP) domain, PETL techniques include inserting lightweight adapter layers~\cite{houlsby2019adapter,bapna2019simple,karimi2021compacter}, learning low-rank adaptation of tunable weights~\cite{hu2021lora}, and prepending learnable prompts to the input sequence~\cite{lester2021power,gu2021ppt,li2021prefix}.
PETL has demonstrated the ability to achieve comparable or even better performance than full fine-tuning while significantly reducing the number of trainable parameters.
This success has inspired extensive studies in computer vision~\cite{zhu2023visual,jia2022visual,gao2023compositional,park2024fair,yang2024fine,wang2024learning} and vision-language~\cite{chen2022plot,huang2023vop,khattak2023maple,zhou2022conditional,roy2023consistency,khattak2023self,zhang2024dept,yao2024tcp} domains. 

In the field of audiovisual learning, Lin~\etal introduced LAVisH~\cite{lin2023lavish}, which transfers knowledge from pretrained vision transformers (\eg, ViT~\cite{dosovitskiy2020vit} and Swin-Transformer~\cite{liu2022swinv2}) to downstream audiovisual tasks. 
LAVisH keeps vision transformers frozen and fine-tunes only cross-modal adapters inserted into each transformer layer.
Subsequent works have explored improved cross-modal interactions using more advanced adapters~\cite{duan2024dgsct,cheng2024avmoe,wang2024stgcma} or prompts~\cite{mahmud2024maavt,zhao2024CoPL}.
For example, DG-SCT~\cite{duan2024dgsct} proposes a dual-guided spatial-channel-temporal attention mechanism to enhance audiovisual feature encoding within transformer layers; AVMoE~\cite{cheng2024avmoe} employs a mixture-of-experts strategy to dynamically adjust uni-modal and cross-modal adapters. 
These studies also leverage modality-specific pretrained transformers (\eg, Swin-V2-L~\cite{liu2022swinv2} for visual and HTS-AT~\cite{chen2022htsat} for audio), leading to improved performance.
However, despite these advancements, these PETL-based methods incur significant memory overhead, as training the inserted adapters or prompts requires gradient propagation back through deep transformer layers.

\subsection{Memory-Efficient Transfer Learning}
Memory-Efficient Transfer Learning (METL) 
is still in its early stages, with most studies primarily focusing on the vision or NLP domains.
Early works aim to reduce training memory through computational optimizations, such as mixed precision~\cite{micikevicius2017mixed}, quantization strategies~\cite{wang2018training}, or by avoiding activation storage~\cite{gomez2017reversible}.
More recently, LST~\cite{sung2022lst} proposes training a lightweight auxiliary network separate from the pretrained transformer, restricting gradient flow to the smaller network.
Other approaches model interactions across transformer layers using lightweight layers~\cite{diao2024unipt} or limit gradient backpropagation to only the final transformer layer~\cite{diao2024sherl}.
However, these methods primarily focus on METL in the vision-language domain and have not been evaluated on the audiovisual downstream tasks.
Moreover, they are not compatible with the pixel-level segmentation task studied in this paper. 
Technically, our method differs in that it learns and utilizes compact meta-tokens for downstream adaptation and is evaluated on both audiovisual classification and segmentation tasks.
Our experiments also indicate that our method achieves a better trade-off between accuracy and efficiency compared to these methods.

\begin{figure*}[t]
  \centering
\includegraphics[width=1\textwidth]{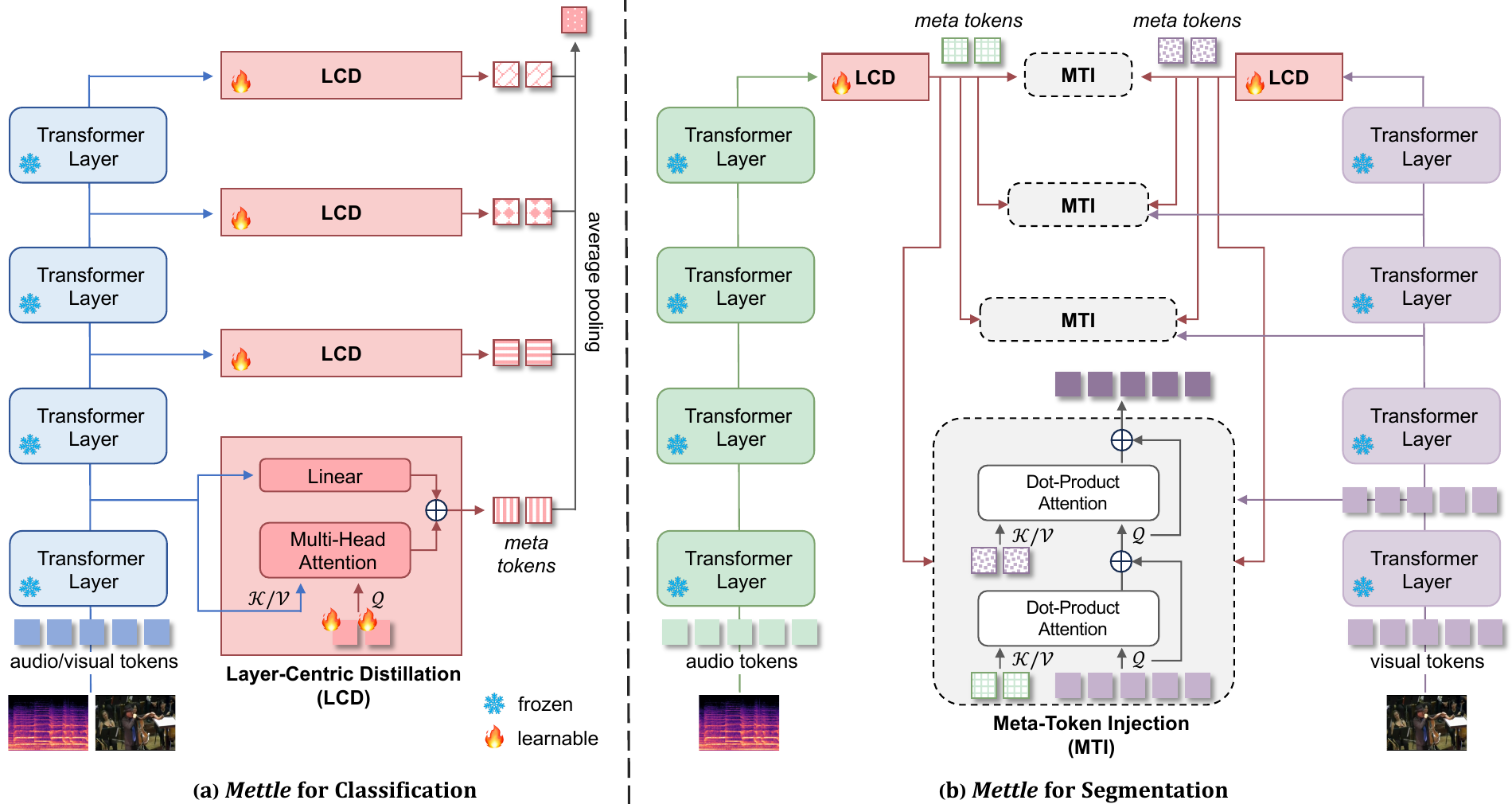}
   \caption{\textbf{Illustration of our Meta-Token Learning ({Mettle}) framework.} We introduce {\textit{Mettle}} for both classification tasks, including AVEL~\cite{tian2018audio} and AVVP~\cite{tian2020unified} and the segmentation task, namely AVS~\cite{zhou2022avs}. 
   \textbf{(a)} For the classification task, we propose the \textit{Layer-Centric Distillation (LCD)} module to distill features from each pretrained transformer layer into learnable meta-tokens. The distilled meta-tokens across transformer layers are further aggregated using simple average pooling for class prediction. LCD is applied at each timestamp for audio and visual modalities. 
   \textbf{(b)} For the segmentation task, LCD is applied only at the final stage of the pretrained transformers to capture high-level semantics. Then, through the \textit{Meta-Token Injection (MTI)} module, the distilled audio and visual meta-tokens are injected into fine-grained visual tokens embedded by earlier transformer layers, better adapting the features for downstream task.
   }
   \label{fig:framework}
\end{figure*}

\section{Method}\label{sec:method}
In this section, we elaborate on the proposed Meta-Token Learning ({\textit{Mettle}}), a memory-efficient approach for adapting large-scale pretrained transformers to downstream audiovisual tasks.
An overview of {\textit{Mettle}} is illustrated in Figure~\ref{fig:framework}.
We first introduce the \textit{Layer-Centric Distillation (LCD)} module, which distills {meta-tokens} from representations embedded by each pretrained transformer layer.
LCD proves effective for audiovisual classification tasks, including AVEL~\cite{tian2018audio} and AVVP~\cite{tian2020unified}.
To extend \textit{Mettle} to the pixel-level segmentation task, AVS~\cite{zhou2022avs}, we introduce the \textit{Meta-Token Injection (MTI)} module.
Both LCD and MTI are designed with parameter efficiency in mind, making our method both parameter and memory efficient. 
The following subsections provide a detailed explanation of our method.

\subsection{Audio-Visual Tokenization}
Given a $T$-second audible video, we extract an RGB frame and a corresponding 1-second mel-spectrogram at each timestamp. Both modalities are tokenized using a patch-based embedding strategy, where the inputs are divided into non-overlapping patches of fixed size. A linear~\cite{dosovitskiy2020vit,liu2022swinv2} or CNN-based~\cite{chen2022htsat} embedding layer is then applied to obtain initial visual token features $\bm{v}_t \in \mathbb{R}^{N_v \times D_v}$ and audio token features $\bm{a}_t \in \mathbb{R}^{N_a \times D_a}$, where $N_v$ and $N_a$ denote the number of tokens, and $D_v$ and $D_a$ represent the feature dimensions. These tokenized features are then fed into $L$-layer pretrained transformers~\cite{dosovitskiy2020vit,liu2022swinv2,chen2022htsat} for feature embedding.

\subsection{\textbf{{\textit{Mettle}}} for Classification}\label{sec:mtl_cls}

\noindent\textbf{Layer-Centric Distillation (LCD).}
Studies have shown that pretrained transformer features encode rich information, with different layers capturing distinct patterns~\cite{ghiasi2022vision,zhang2022layerscreatedequal,becker2018interpreting}.
The Layer-Centric Distillation (LCD) module is proposed to extract essential knowledge from each transformer layer, condensing it into \textit{meta-tokens} for downstream tasks. As illustrated in Figure~\ref{fig:framework}a, LCD operates independently for the audio and visual modalities.
Next, we describe the LCD process using the visual modality as an example.

Given the visual tokens $\bm{v}_t^l \in \mathbb{R}^{N_v^l \times D_v^l}$ from the $l$-th layer of a pretrained transformer ($l=\{1,2,...,L\}$), the meta-tokens $\bm{m}_t^l \in \mathbb{R}^{K \times D_{v}^l}$ are \textit{task-specific} learnable vectors with the same channel dimension as $\bm{v}_t^l$. These meta-tokens are randomly initialized and refined during later distillation. $K$ is the number of meta-tokens, which may differ between audio and visual modalities, and $K \ll N_v^l$.

To distill $\bm{v}_t^l$ into $\bm{m}_t^l$, we leverage the multi-head attention~\cite{vaswani2017attention} mechanism.
Specifically, the meta-tokens $\bm{m}_t^l$ serve as the \textit{query}, while the visual tokens $\bm{v}_t^l$ act as \textit{key} and \textit{value}, formulated as: 
\begin{equation}
\begin{aligned}
\label{eq:lcd_mha}
\bm{m}^l_t \leftarrow \mathrm{Softmax}\left( (\bm{m}^l_t \bm{W}^l_{\mathcal{Q}})(\bm{v}_t^l \bm{W}^l_{\mathcal{K}})^\top\right)(\bm{v}^l_t \bm{W}^l_{\mathcal{V}}),
\end{aligned}
\end{equation}
where $\bm{W}^l_{\mathcal{Q}}$,$\bm{W}^l_{\mathcal{K}}$,$\bm{W}^l_{\mathcal{V}} \in \mathbb{R}^{D_v^l \times D_v^l}$ are learnable parameters. The operator `$\leftarrow$' indicates that the meta-tokens $\bm{m}_t^l$ are updated. By capturing the interactions between $\bm{m}^l_t$ and $\bm{v}^l_t$ through dynamic attention weights, task-relevant representations from $\bm{v}^l_t$ are effectively distilled into the more compact $\bm{m}^l_t$, enhancing their \textit{task-specific} adaptability.

Meanwhile, we introduce a parallel pathway that performs direct token reduction via a linear projection, providing a complementary representation that maintains a more direct relationship with the original \textit{pretrained knowledge}.

Consequently, the final distilled meta-tokens $\bm{m}_t^l$ are computed as:
\begin{equation}
\begin{aligned}
\label{eq:lcd_global}
\bm{m}^l_t \leftarrow \bm{m}_t^l + \bm{W}^l_{g} \bm{v}^l_t,
\end{aligned}
\end{equation}
where $\bm{W}^l_{g} \in \mathbb{R}^{K \times N_v^l}$ is the learnable parameter of linear layer.
This process can be repeated for $R$ iterations, forming an $R$-step layer-centric distillation.

As shown in Figure~\ref{fig:framework}a, LCD is applied to each pretrained transformer layer, generating layer-wise meta-tokens $\bm{m}^l_t \in \mathbb{R}^{K \times D_v^l} (l=\{1,...,L\})$.
In hierarchical transformers such as Swin-V2-L~\cite{liu2022swinv2}, the channel dimension $D_v^l$ varies across layers. To ensure consistency, 
an independent linear layer is applied to project each layer's meta-token dimension to a fixed value $d_v$.
Then, meta-tokens from multiple layers are aggregated into a single meta-token $\bm{\overline{m}}_t \in \mathbb{R}^{1\times d_v}$ using parameter-free average pooling, formulated as,
\begin{equation}
\begin{aligned}
\bm{\overline{m}}_t = \frac{1}{LK} \sum_{l=1}^{L}\sum_{k=1}^{K} \bm{m}_{t,k}^l,
\end{aligned}
\end{equation}
where $\bm{m}_{t,k}^l$ is the $k$-th token of $\bm{m}_{t}^l$ ($k=\{1,...,K\}$). 
Such cross-layer aggregation allows the model to comprehensively utilize diverse and useful information embedded in meta-tokens across multiple layers.

By extending the above LCD operations for all $T$ 
timestamps of visual and audio modalities, we obtain the segment-level visual and audio meta-tokens, denoted as $\bm{\overline{v}} \in \mathbb{R}^{T\times d_v}$ and $\bm{\overline{a}} \in \mathbb{R}^{T\times d_a}$, respectively.
Then, they can be utilized for downstream audiovisual classification tasks (\eg, AVEL~\cite{tian2018audio} and AVVP~\cite{tian2020unified}) following prior works~\cite{duan2024dgsct,cheng2024avmoe}.

\subsection{\textbf{{\textit{Mettle}}} for Segmentation}\label{sec:mtl_seg}
For segmentation tasks such as AVS~\cite{zhou2022avs}, compact meta-tokens alone are insufficient for generating pixel-level segmentation maps, which typically rely on multiple high-resolution visual feature maps encoded by pretrained transformers~\cite{duan2024dgsct,cheng2024avmoe}.
Therefore, we consider enhancing the multi-resolution visual features using the distilled audio and visual meta-tokens.
We explain the details next.

\noindent\textbf{Layer-Centric Distillation (LCD).}
We apply LCD exclusively to the final layer of the pretrained transformer for meta-token distillation, as features embedded from the later stages of deep neural networks capture higher-level global semantics~\cite{ghiasi2022vision} (we will provide more discussions on this design in Sec.~\ref{sec:ablation_MTI}).
This is particularly relevant in the Swin-V2-L~\cite{liu2022swinv2} backbone, where the final stage has a larger receptive field.
The meta-token distillation process follows Eqs.~\ref{eq:lcd_mha} and~\ref{eq:lcd_global}.
The distilled audio and visual meta-tokens at the $t$-th timestamp are denoted as $\bm{m}_t^a \in \mathbb{R}^{K_a \times d_a}$ and $\bm{m}_t^v \in \mathbb{R}^{K_v \times d_v}$, respectively, where $d_a = d_v$.

\noindent\textbf{Meta-Token Injection (MTI).}
Unlike the direct aggregation of meta-tokens used in classification tasks,
MTI injects knowledge encoded in meta-tokens $\bm{m}_t^a$ and $\bm{m}_t^v$ into multi-resolution visual features extracted by pretrained transformer layers, adapting them for downstream AVS task, \ie, improving their ability to capture sounding objects.

Specifically, given the visual tokens $\bm{v}_t^l \in \mathbb{R}^{N_v^l \times D_v^l}$ extracted by the $l$-th transformer layer, the audio and visual meta-tokens are first processed by independent linear layers to match the channel dimensions of $\bm{v}_t^l$.
Then, MTI consists of two steps: \textit{cross-modal} injection and \textit{intra-modal} injection. As shown in Figure~\ref{fig:framework}b, the aligned audio meta-token $\bm{m}_t^a \in \mathbb{R}^{K_a \times D_v^l}$ and visual meta-token $\bm{m}_t^v \in \mathbb{R}^{K_v \times D_v^l}$ sequentially interact with the visual tokens $\bm{v}_t^l$.
In each injection step, dot-product attention~\cite{vaswani2017attention} is first used to encode the cross-modal or intra-modal relations between meta-tokens and visual tokens, followed by a residual connection.
This process is formulated as follows:
\begin{equation}
\begin{aligned}
\label{eq:mti}
\bm{v}^l_t \leftarrow \bm{v}^l_t + \mathrm{Softmax}\left( \bm{v}^l_t (\bm{m}^a_t)^\top \right) \bm{m}^a_t,\\
\bm{v}^l_t \leftarrow \bm{v}^l_t + \mathrm{Softmax}\left( \bm{v}^l_t (\bm{m}^v_t)^\top \right) \bm{m}^v_t.
\end{aligned}
\end{equation}

Apart from several preprocessing linear layers, the above injection process is parameter-free, 
distinguishing it from prior methods that employ complex audio-visual adapters with a large number of learnable parameters~\cite{lin2023lavish,duan2024dgsct,cheng2024avmoe,wang2024stgcma}.
Moreover, the MTI is inserted only into the four stages of the pretrained transformer. The updated visual tokens $\bm{v}^l_t$ from these stages are sent to an FPN-based decoder~\cite{zhou2022avs} in the AVS task to generate segmentation maps.

\section{Experiments}\label{sec:experiments}

\input{./tables/sota_ave}

\subsection{Experimental Setups}\label{sec:exp_setup}
\noindent\textbf{Datasets and Metrics.}
We evaluate our method on three downstream audio-visual learning tasks: Audio-Visual Event Localization ({AVEL}), Audio-Visual Video Parsing ({AVVP}), and Audio-Visual Segmentation ({AVS}). The definitions of these tasks are provided in Sec.~\ref{sec:related_work_avl}. Here, we briefly introduce the corresponding datasets and evaluation metrics.
For the \textbf{AVEL} task, we conduct experiments on the AVE~\cite{tian2018audio} dataset, which contains 4,143 videos across 28 event categories. 
Following prior works, we use segment-level accuracy (Acc.) to evaluate predicted audio-visual events.
For the \textbf{AVVP} task, we use the LLP~\cite{tian2020unified} dataset, which consists of 11,849 videos spanning 25 categories.
Following~\cite{tian2020unified}, we report the F1-score of predicted audio events, visual events, and audio-visual events at both segment-level and event-level.
For the \textbf{AVS} task, we evaluate both single-source and multi-source segmentation using the S4 and MS3 subsets of AVSBench~\cite{zhou2022avs}. These subsets contain 4,932 and 404 videos, respectively, covering 23 classes. We use mean Intersection over Union (mIoU) as the evaluation metric.

\noindent\textbf{Implementation Details.} We train and evaluate our model using various large-scale pretrained transformer backbones, including ViT-L-16~\cite{xu2022groupvit}, Swin-V2-L~\cite{liu2022swinv2}, and HTS-AT~\cite{chen2022htsat}.
In some experiments, following prior work LAVisH~\cite{lin2023lavish}, we use ViT-L-16 or Swin-V2-L for both audio and visual modalities (\textit{shared}) to show that a model pretrained only in the vision domain can be adapted to audiovisual domain.
As illustrated in Figure~\ref{fig:framework}, our {\textit{Mettle}} approach serves as the early audio and visual encoder. Task-specific decoders are then applied to generate predictions following~\cite{duan2024dgsct,cheng2024avmoe}.
Additional details on batch size, learning rate, training epochs, and other training hyper-parameters for different tasks are provided in Sec.~\ref{sec:supp_training_config} of our Appendix.
All of our experiments are conducted on an NVIDIA A100-SXM4-40GB GPU.

\subsection{Main Results}\label{sec:main_results}
In this section, we present quantitative comparisons of our \textit{Mettle} method against prior approaches on downstream audiovisual tasks, with a particular focus on recent parameter-efficient tuning methods, \eg, LAVisH~\cite{lin2023lavish}, DG-SCT~\cite{duan2024dgsct}, and AVMoE~\cite{cheng2024avmoe}.
For each task, in addition to the task-specific evaluation metrics introduced in Sec.~\ref{sec:exp_setup}, we report the proportion of trainable parameters (\%) and the total number of parameters (M) to assess parameter efficiency.
Additionally, we measure training memory usage (GB) and runtime (ms)\footnote{We do not adopt the training GPU hours as the metric, as it is hard to fairly determine the optimal convergence points of prior methods during our reproduction. But, the GPU hours are expected to follow similar trends as the runtime.} to evaluate memory efficiency.
Notably, these two metrics, ignored by prior works, are recorded during the model training process and normalized by batch size.

\input{./tables/sota_avvp}

\subsubsection{Results on Audio-Visual Event Localization}
Table~\ref{tab:sota_ave} presents the comparison results.
We evaluate our method using various transformer backbones, including modality-shared ViT-L-16~\cite{dosovitskiy2020vit} and Swin-V2-L~\cite{liu2022swinv2}, as well as modality-separate Swin-V2-L (visual) and HTS-AT~\cite{chen2022htsat} (audio).
As shown in the Table, our method achieves comparable or superior accuracy compared to prior methods.
Notably, our approach remains effective even with modality-shared ViT-L-16 and Swin-V2-L backbones, demonstrating that the vision-domain knowledge can be memory-efficiently adapted to downstream audiovisual tasks.
This also highlights the generalization capability of our method.
The last row of Table~\ref{tab:sota_ave} shows that our method achieves the highest performance when using modality-separate backbones.
This is expected, as HTS-AT is pretrained on an audio-domain dataset AudioSet~\cite{gemmeke2017audioset}, enabling it to encode more effective audio representations, thereby improving meta-token distillation.
More importantly, our approach achieves competitive accuracy with significantly lower parameters and memory costs.
For instance, compared to the state-of-the-art method AVMoE~\cite{cheng2024avmoe}, our model reduces the total parameter count by 66M and decreases the proportion of learnable parameters by 11.9\%.
Moreover, AVMoE requires 12.1GB of memory per sample and 256.1ms per iteration during training.
In contrast, our model requires only 1.5GB and takes 88.3ms, demonstrating superior efficiency.

\subsubsection{Results on Audio-Visual Video Parsing}
We compare our method with prior models on the audio-visual video parsing task.
For a fair comparison with prior works~\cite{duan2024dgsct,cheng2024avmoe}, we adopt Swin-V2-L~\cite{liu2022swinv2} and HTS-AT~\cite{chen2022htsat} as the visual and audio transformer backbones, respectively.
As shown in Table~\ref{tab:sota_avvp}, our method outperforms DG-SCT~\cite{duan2024dgsct} by 1.1\% in the average metric (`Avg.') and achieves competitive performance with the state-of-the-art AVMoE~\cite{cheng2024avmoe}.
We observe an imbalance between audio prediction (`A') and visual prediction (`V') metrics. This discrepancy arises from the inherent challenges of the AVVP task, where only weak video-level labels are available, and the modality-specific event labels are unknown.
To ensure a fair comparison with recent parameter-efficient methods, we do not utilize additional video-level or segment-level pseudo labels proposed in recent studies~\cite{wu2021exploring,cheng2022joint,lai2023modality}.
Despite this constraint, our method maintains competitive accuracy under the same conditions while demonstrating superior parameter and memory efficiency.
Compared to AVMoE~\cite{cheng2024avmoe}, our \textit{Mettle} significantly reduces the trainable parameters (30.9\% $\rightarrow$ {8.3\%}) and total parameters (376.4M $\rightarrow$ {283.5M}). Moreover, our model dramatically lowers memory consumption (12.9GB $\rightarrow$ {0.6GB}) and training runtime (490.0ms $\rightarrow$ {64.2ms}).

\input{./tables/sota_avs}

\subsubsection{Results on Audio-Visual Segmentation}
Table~\ref{tab:sota_avs} shows the comparison results for the pixel-level AVS task.
Our method demonstrates superior accuracy, as well as improved parameter and memory efficiency. 
When using the modality-shared Swin-V2-L backbone, our approach attains comparable mIoU to LAVisH~\cite{lin2023lavish} in the single-source S4 setting and significantly outperforms it by 3.9\% mIoU in the multi-source MS3 setting. Furthermore, our model requires fewer parameters and consumes only half the training memory and runtime of LAVisH.
The efficiency gains become even more pronounced when employing modality-separate backbones. As shown in the lower part of Table~\ref{tab:sota_avs}, compared to prior method AVMoE~\cite{cheng2024avmoe}, our method achieves comparable mIoU in the S4 setting and surpasses prior work in the MS3 setting, while reducing trainable parameter count by 77\%, memory by 69\%, and runtime by 75\%.

{\subsection{Further Comparisons}}\label{sec:further_comparison}

\noindent In Sec.~\ref{sec:main_results}, we presented extensive experiments comparing our method with existing parameter-efficient audiovisual fine-tuning methods~\cite{lin2023lavish,duan2024dgsct,cheng2024avmoe}, demonstrating its superiority in parameter efficiency, memory usage, and training time while maintaining competitive performance.
In this section, we provide additional comparisons with several representative memory-efficient methods from the vision-language domain, as well as two variant baseline approaches.
The details are described below.

\input{./tables/tab_compare_with_other_memory_efficient_methods}

\subsubsection{Comparison with Memory-efficient Methods from the Vision-language field}

Current exploration of the memory-efficient methods primarily focus on the vision-language domain, while our work represents the first attempt in the audiovisual field.
For a comprehensive comparison, we try to adapt several state-of-the-art memory-efficient methods from the vision-language domain -- LST~\cite{sung2022lst}, UniPT~\cite{diao2024unipt}, and SHERL~\cite{diao2024sherl}, to the studied audiovisual downstream tasks.
As these methods are not applicable to pixel-level segmentation, we focus our comparisons on the AVEL and AVVP classification tasks.
As shown in Table~\ref{tab:comparison_with_memory_efficient_methods}, our method uses fewer parameters, and achieves comparable training memory usage and runtime compared with these meticulously designed baselines. 
Moreover, our method consistently outperforms them in accuracy on both tasks, demonstrating a better trade-off between accuracy and efficiency.

\subsubsection{Comparison with two vanilla Audiovisual Approaches}
We also compare our method with two vanilla approaches: 1) \textit{Direct Connection}. Audio and visual features extracted by the pretrained transformer backbones are directly used for downstream tasks.
2) \textit{Linear Probing}. A modality-independent linear layer is added to the final layer of the pretrained transformer backbone for task-specific adaptation. Notably, for the AVS~\cite{zhou2022avs} task, the features updated by linear probing are further processed by our proposed meta-token injection (MTI) modules.
Table~\ref{tab:comparison_with_linearprobing} presents the comparison results.
Compared to the \textit{Direct Connection} approach, although the \textit{Linear Probing} strategy introduces learnable layers, it does not improve performance on the AVEL and AVVP tasks but proves beneficial for AVS when combined with our MTI module.
Nevertheless, our method, which uses layer-centric distillation and meta-token injection, significantly outperforms both approaches.

\input{./tables/supp_tab_compare_with_frozen_linearprob}

\subsection{Ablation Studies}\label{sec:ablations}
In this section, we conduct ablation studies to analyze the effects of key hyper-parameters and the core components of our Layer-Centric Distillation (\textbf{LCD}) and Meta-Token Injection (\textbf{MTI}) modules.
{For simplicity, we primarily report the accuracy-based metrics in the ablations.}
Unless otherwise specified, we use a modality-shared Swin-V2-L~\cite{liu2022swinv2} backbone for the AVEL and AVS tasks.
For the AVVP task, modality-specific backbones are adopted, with Swin-V2-L for the visual modality and HTS-AT~\cite{chen2022htsat} for the audio modality.

\input{./tables/Ab_meta_token_and_step_num}

\begin{figure}[t]
  \centering
  \includegraphics[width=0.4\textwidth]{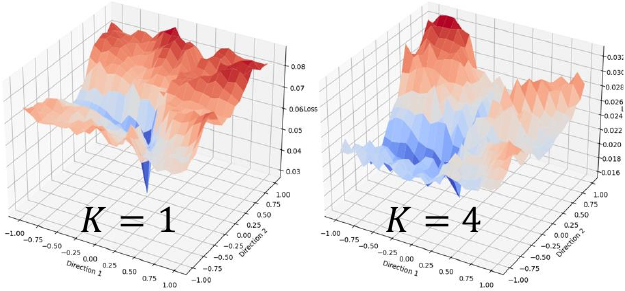}
  \caption{
  \textbf{Loss landscape of models trained with different numbers of meta-tokens.} The subfigures are generated using models with a modality-shared Swin-V2-L backbone, evaluated on the AVEL task.}
  \label{fig:loss_landscape}
\end{figure}

\subsubsection{Hyperparameter Studies}

\noindent\textbf{Effect of the Meta-Token Number $K$.}
In the LCD module, we introduce $K$ learnable meta-tokens distilled from pretrained transformer layers. The numbers may be different for audio and visual modalities, denoted as $K_a$ and $K_v$, respectively.
Table~\ref{tab:meta_token_number} presents the ablation results.
For most downstream audiovisual tasks, using a single meta-token per modality seems sufficient to achieve high performance.
This design ensures that the learned meta-token remains highly compact for downstream adaptation while minimizing redundancy in features encoded by pretrained backbones.
{To provide a more in-depth analysis, in Figure~\ref{fig:loss_landscape}, we plot the loss landscapes of models trained with 
different numbers of meta tokens ($K=\{1,4\}$), evaluated on the AVEL task.
The comparison reveals that using a larger number of meta tokens results in a more rugged optimization landscape with multiple local minima, which may hinder convergence. In contrast, fewer meta tokens lead to a smoother landscape, potentially making the model easier to optimize and improving performance.}
The optimal token number may differ slightly when using the modality-separate backbones (Swin-V2-L \& HTS-AT) for AVEL and AVS tasks.
For example, in this setup, the model has the best performance with one audio meta-token and two visual meta-tokens (still quiet few) for the AVEL task.
The final choice of meta-token numbers is provided in Sec.~\ref{sec:supp_training_config} of our Appendix.

\input{./tables/supp_hierarical_metatoken_numers}

\noindent\textbf{Effect of using Hierarchical Design for Meta-Tokens.}
In our LCD module, the audio/visual meta-token numbers $K$ remain identical across different transformer layers (\ie, $K_a=K_v=K$ ).
For transformers with hierarchical designs, \eg, Swin-V2-L~\cite{liu2022swinv2} and HTS-AT~\cite{chen2022htsat},  earlier layers encode more audio/visual tokens, while the later (top) layers contain fewer tokens.
A natural question arises: what if we also adopt a hierarchical number of meta-tokens in the layer-centric distillation process?
To explore this, we test two designs. 
Both Swin-V2-L and HTS-AT consist of four transformer stages.
First, based on prior experimental results, we set the meta-token number at the last stage to 1.
In our first design, we define the meta-token numbers using a factor of 2. As a result, the first (bottom) stage uses 8 meta-tokens, the second stage uses 4 meta-tokens,  and the third stage uses 2, denoted as `\{8,4,2,1\}' in Table~\ref{tab:hierarchical_metatoken_numers}.
In our second design, the top two stages use 1 meta-token each, while the bottom two stages use 2 meta-tokens each, resulting in `\{2,2,1,1\}' in Table~\ref{tab:hierarchical_metatoken_numers}.
As shown in the Table, neither hierarchical design improves the model's performance, indicating that hierarchical meta-token numbers are unnecessary for distillation.
Consequently, we choose to use identical meta-tokens across all transformer layers/stages. This also simplifies the implementation and further reduces the number of learnable parameters.

\input{./tables/Ab_distill_step_number}

\input{./tables/Ab_component_of_lcd_mti}

\noindent\textbf{Effect of the Distillation Step $R$.}
The distillation process operates through $R$-step cycles (Eq.~\ref{eq:lcd_global}), iteratively updating the meta-tokens. We denote the distillation steps for audio and visual modalities as $R_a$ and $R_v$, respectively.
Table~\ref{tab:distill_step_number} presents the ablation results.
A single-step distillation per modality has guaranteed high performances for most downstream tasks.
This design choice keeps our model lightweight, reducing the number of learnable parameters and computational overhead.
By analyzing Table~\ref{tab:meta_token_number} and Table~\ref{tab:distill_step_number}, we observe that classification tasks such as AVEL and AVVP are more sensitive to distillation-related hyperparameters. This is reasonable since these tasks heavily rely on distilled meta-tokens for final classification. In contrast, for the segmentation task, visual feature maps embedded by pretrained transformers play a more critical role in determining final segmentation results.

\begin{figure*}[t]
  \centering
\includegraphics[width=1\textwidth]{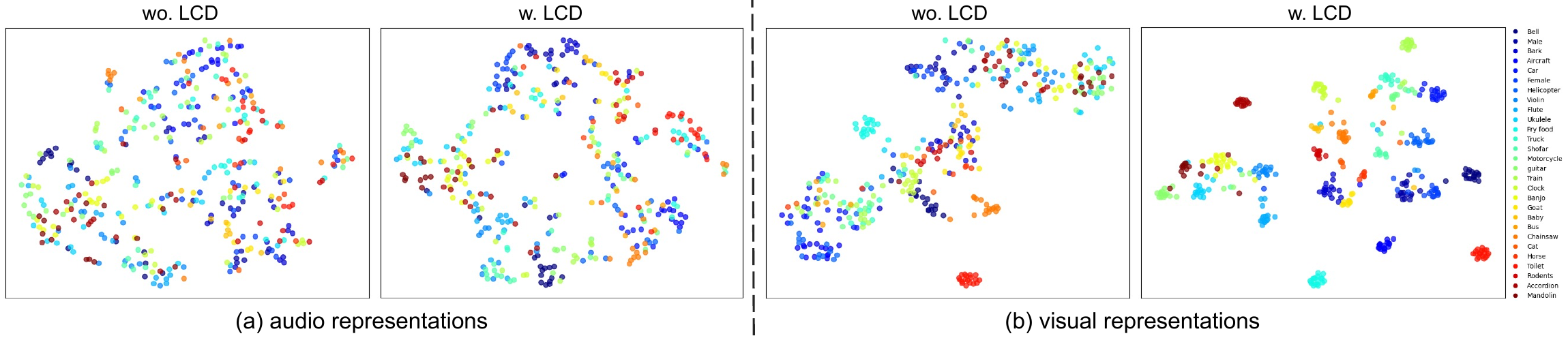}
   \caption{\textbf{T-SNE visualization of audio and visual meta-token representations.} 
   Samples are taken from the test set of the AVE~\cite{tian2018audio} dataset.
   }
   \label{fig:tsne_ave}
\end{figure*}

\subsubsection{Ablation Studies on LCD}

\noindent\textbf{Component Ablation of LCD.}
The LCD module comprises two key components: 1) The learnable meta-tokens are introduced to interact with pretrained features for Task-Specific adaptation (Eq.~\ref{eq:lcd_mha}); 2) A lightweight linear layer is used to better retain Domain-Pretrained knowledge (Eq.~\ref{eq:lcd_global}).
We denote these components as \textbf{TS} and \textbf{DP}, respectively.
The ablation results are shown in Table~\ref{tab:component_for_LCD}.
Removing either component leads to a significant performance drop across all downstream audio-visual tasks.
For example, in the AVEL task, excluding the TS component reduces accuracy from 82.3\% to 77.3\%. These results highlight the necessity of both pretrained knowledge preservation and task-specific adaptation in the LCD module.

\input{./tables/supp_using_avgpooling_branch_in_LCD}

\input{./tables/Ab_component_of_mti}

\noindent\textbf{Linear layer vs. Average Pooling in LCD.}
Our LCD module incorporates a linear layer branch to process the original audio or visual tokens embedded by the pretrained transformer model, complementing the branch that utilizes meta-tokens.
We also explore a parameter-free variant that replaces the linear layer with direct average pooling on the original audio/visual token features.
As shown in Table~\ref{tab:using_avgpooling_in_LCD}, compared to removing the branch (`Remove'), this simple strategy (`AvgPool.') also benefits the AVEL and AVS tasks, highlighting the importance of designing a branch to preserve pretrained knowledge.
Moreover, the average pooling strategy can achieve comparable performance to the learnable linear layer in AVEL and single-source S4 setting of AVS tasks.
The video data in these tasks typically contains a single event or sounding source, making simple average pooling effective.
However, for more complex tasks (\ie, the AVVP and AVS-MS3), which involve multiple events, the linear layer proves more effective for LCD.

\begin{figure*}[t]
  \centering
\includegraphics[width=0.9\textwidth]{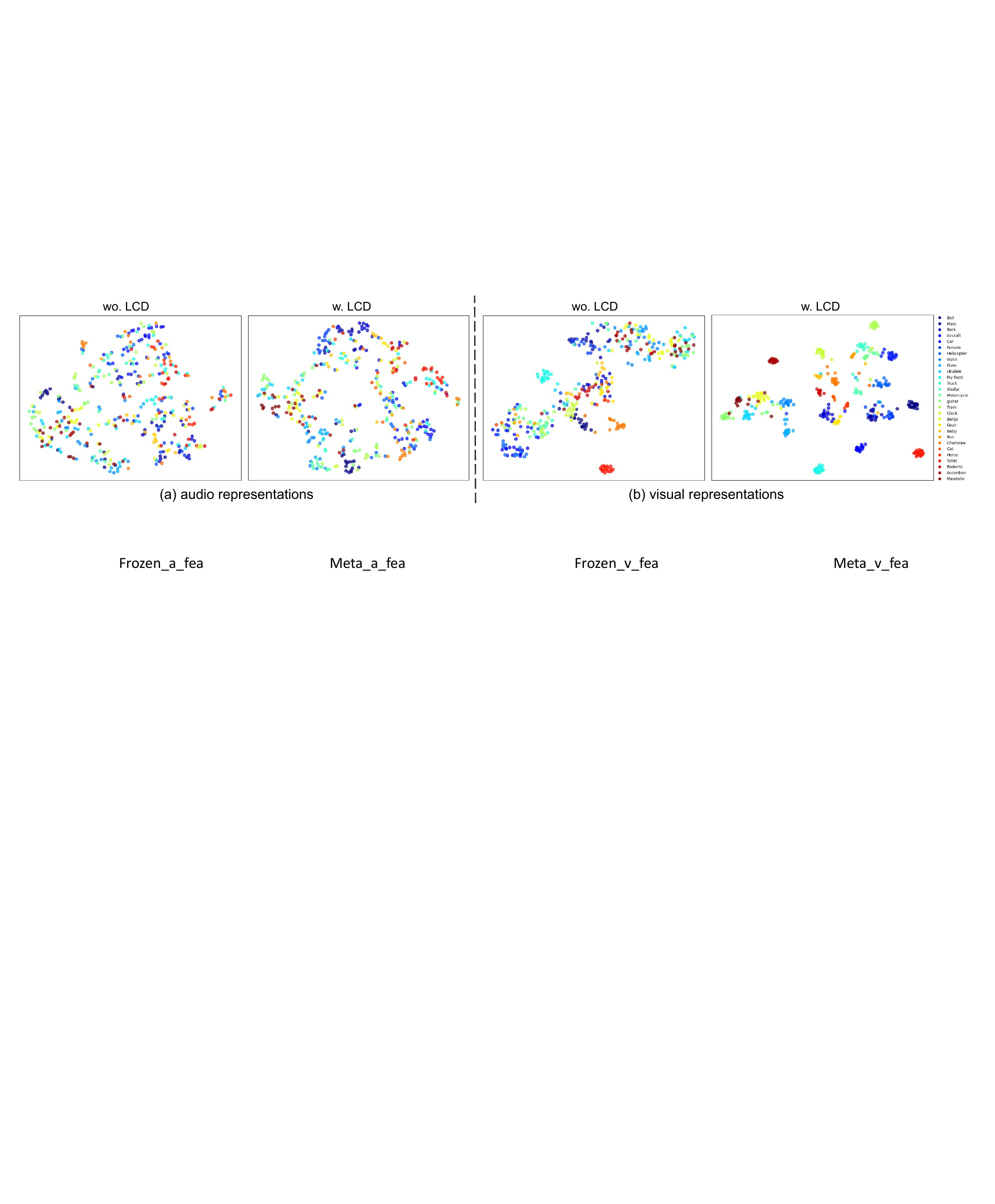}
   \caption{\textbf{Qualitative results for the AVVP task.} The model trained with the LCD module provides more accurate audio and visual event parsing along the timeline.}
   \label{fig:vis_avvp_results}
\end{figure*}

\begin{figure*}[t]
  \centering
\includegraphics[width=0.95\textwidth]{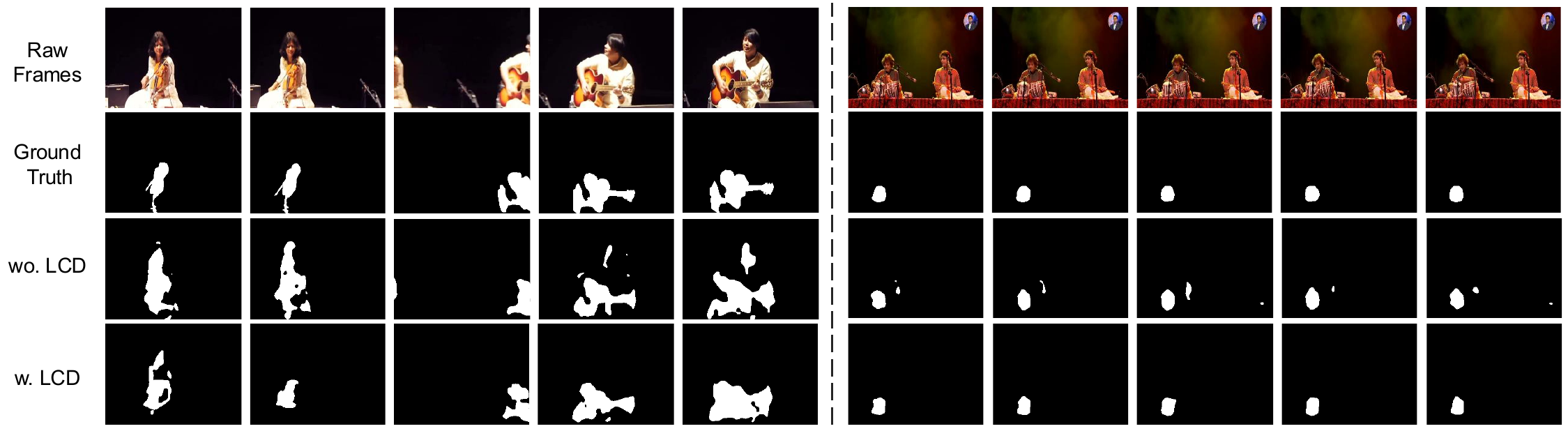}
   \caption{\textbf{Qualitative results for the AVS task.} The model trained with the LCD module generates more accurate segmentation boundaries of the sounding objects.
   }
   \label{fig:vis_avs_results}
\end{figure*}

\subsubsection{Ablation Studies on MTI}\label{sec:ablation_MTI}

\noindent\textbf{Component Ablation of MTI.}
The MTI module is designed for the segmentation AVS task, incorporating both cross-modal and intra-modal injection by utilizing meta-tokens from audio and visual modalities.
We ablate their impacts in Table~\ref{tab:component_for_MTI}.
Removing either intra-modal or cross-modal injection degrades the model's performance to varying degrees, with a more pronounced impact in the more challenging MS3 setting.
For example, the mIoU significantly drops by 3.8\% without using the cross-modal injection.
The MS3 setting involves multiple sound sources, making the guidance from audio meta-tokens crucial for accurately identifying the corresponding sounding objects.

\noindent\textbf{Injection order in MTI module.}
Our MTI module first applies cross-modal injection followed by intra-modal injection  (`cross-first').
We explore the impact of reversing this order, \ie, the intra-modal injection first.
In this setup, the mIoU for the S4 and MS3 settings of the AVS task is 79.8\% and 53.3\%, respectively, which is very close to the `cross-first' setup (79.9\% and 53.7\%).
This suggests that the injection order has minimal impact on the model's performance.

\noindent\textbf{Placement of the MTI Module.}
As shown in Figure~\ref{fig:framework}b, the MTI module is applied by default to combine with the LCD-distilled meta tokens extracted from the last (top) pretrained transformer layer.
This design is based on the hypothesis that meta tokens from higher layers contain more accurate and comprehensive global category semantics.
To validate this, we conduct experiments on the AVEL classification task by applying LCD to individual transformer layers (1st to 4th). The resulting accuracies are 49.5\%, 64.7\%, 80.4\%, and 81.7\%, respectively.
This trend suggests that top-layer features indeed capture richer category-level semantics, justifying their use for guiding the MTI module.
Furthermore, we apply LCD to both the final and earlier layers and feed the resulting meta tokens into the MTI module.
As shown in Table~\ref{tab:MTI_placement}, MTI benefits more from high-layer LCD, particularly in the more challenging MS3 setting.

\begin{table}[t]
    \belowrulesep=0pt
    \aboverulesep=0pt
\caption{\textbf{Effect of MTI module placement along the transformer depth.} Experiments are conducted on the AVS task, and the mIoU metric is reported.}
\centering
\tabcolsep 4pt
\begin{tabular}{c|cccc}
\hline
     layer index & 1st (bottom)  & 2nd & 3rd & 4th (top) \\
    \midrule
    S4 & 80.2&80.3&80.5&\textbf{80.7}  \\ 
    MS3 &53.6&53.8&54.5&\textbf{55.1} \\
        \hline
\end{tabular}
\label{tab:MTI_placement}
\end{table}

\subsection{Qualitative Results}
The effectiveness of our approach primarily stems from the layer-centric distillation (LCD), which generates compact meta-tokens to facilitate downstream task adaptation.
We provide qualitative results to demonstrate its advantages.
Based on the AVEL task, we visualize the representations of learned audio and visual meta-token representations (`w.~LCD') and compare them to features produced by the frozen pretrained backbone (`wo.~LCD').
The Swin-V2-L is used as shared backbone.
As shown in Figure~\ref{fig:tsne_ave}, the distilled meta-tokens exhibit better intra-class compactness and inter-class separation for both modalities.
These results further validate the effectiveness of our meta-token distillation process: the meta-tokens adapts better to downstream task.
We also present some qualitative results on the AVVP and AVS tasks. 
As shown in Figure~\ref{fig:vis_avvp_results}, our model trained with the layer-centric distillation (`w. LCD') more precisely localizes the temporal segments of events, highlighted by the dashed line box.
In Figure~\ref{fig:vis_avs_results}, it better delineates the shape of sounding objects and reduces the over-segmentation of unrelated pixels.

\vspace{-1ex}
\section{Conclusion and Future Work}
\label{sec:conclusion}
We found that prior parameter-efficient audio-visual adaptation methods, designed to adapt large-scale pretrained transformers to downstream tasks, exhibit significant memory inefficiency.
Our \textit{Mettle} approach mitigates this issue by distilling audio and visual meta-tokens from pretrained backbones in parallel.
Specifically, we introduce lightweight layer-centric distillation and meta-token injection modules. The former module performs layer-wise distillation of audio and visual tokens from pretrained transformers into compact meta-tokens, while the latter injects the extracted top-layer meta-tokens into high-resolution audio and visual tokens from earlier layers.
Evaluated on three audiovisual classification and segmentation tasks, including AVEL, AVVP, and AVS, our method excels in accuracy, parameter efficiency, memory usage, and training speed, achieving a favorable trade-off among these factors.
We will release the source code to the community and hope to inspire further progress in memory-efficient audio-visual adaptation.

\vspace{-1ex}
\noindent\textbf{Limitation.}
Despite its effectiveness, our method faces challenges in more complex \textit{reasoning} tasks like audiovisual question answering~\cite{li2022learning}, which involves diverse question types (\eg, \textit{location}, \textit{counting}, and \textit{comparative}) within long, dynamic audio-visual scenarios.
Effectively distilling compact meta-tokens that generalize across various questions remains difficult relying solely on frozen intermediate features.
To overcome this, our future work will explore integrating memory-efficient modules with partial layer fine-tuning to enhance adaptability to more complex tasks.

\appendix
\onecolumn

\section*{Appendix}\label{sec:appendix}

\section{Training hyperparameters}\label{sec:supp_training_config}

Table~\ref{tab:supp_training_hyperparameters} presents the hyperparameter details for training our model on downstream audiovisual learning tasks, 
including audio-visual event localization (AVEL)~\cite{tian2018audio}, audio-visual video parsing (AVVP)~\cite{tian2020unified}, and audio-visual segmentation (AVS)~\cite{zhou2022avs}.
We use the same training configurations as shown in Table~\ref{tab:supp_training_hyperparameters} to train our model with both the modality-shared (Swin-V2-L~\cite{liu2022swinv2}) and modality-specific (Swin-V2-L for visual, HTS-AT~\cite{chen2022htsat} for audio) transformer backbones.
For the AVEL and AVS tasks, our ablation studies on hyperparameters related to the layer-centric distillation process (\ie, the meta token number and distillation step) presented in Sec.~\ref{sec:ablations} are based on the model trained with modality-shared Swin-V2-L backbone.
The best hyperparameters for using modality-specific backbones are provided in Table~\ref{tab:supp_parameters_for_modality_specific_backbones}.
We will release the source code to facilitate reproducibility and support further advancements in memory-efficient audiovisual adaptation.

\input{./tables/tab_training_configures}
\input{./tables/supp_hyperparameter_for_modality_specific_backbones}

\section{Extending the Method to CNN Backbone}
As demonstrated in the main paper, our method supports hierarchical backbones such as Swin-V2-L~\cite{liu2022swinv2}, and can therefore be extended to CNN-based backbones as well.
In Table~\ref{tab:extend_to_cnns}, we report results using ResNeXt101~\cite{xie2017aggregated} on the AVEL classification and AVS-MS3 segmentation tasks.

\begin{table}[hp]
    \belowrulesep=0pt
    \aboverulesep=0pt
\centering
\caption{\textbf{Results of the model using the ResNeXt101 CNN backbone.}}
\tabcolsep 2.5pt
\begin{tabular}{ccc|c|ccc|c}
\hline
\multicolumn{4}{c|}{AVEL (classification)} & \multicolumn{4}{c}{AVS-MS3 (segmentation)}  \\ \midrule
     Param. (M) & Mem. (G) & Time (ms). & Acc. &  Param. (M) & Mem. (G) & Time (ms). & mIoU\\
    \midrule
     151.6&1.0&41.8&80.1  &151.9&3.7&86.7&48.4 \\
        \hline
\end{tabular}
\label{tab:extend_to_cnns}
\end{table}

\begin{table}[t]
    \belowrulesep=0pt
    \aboverulesep=0pt
\centering
\caption{\textbf{Inference runtime comparison.}}
\scriptsize
\tabcolsep 3pt
\begin{tabular}{c|ccc}
\hline
{Methods} & {AVEL} & {AVVP} & {AVS} \\
\hline
DG-SCT~\cite{duan2024dgsct} & 253.2 & 162.8 & 240.7 \\
AVMoE~\cite{cheng2024avmoe} & 140.5 & 287.5 & 191.6 \\
\textbf{Ours} & \textbf{84.4} & \textbf{60.2} & \textbf{73.7} \\
\hline
\end{tabular}
\label{tab:inference_runtime_comparison}
\end{table}

\section{Inference runtime comparison} 

In the main paper, we compare the \textit{training}-time runtime of our method with prior audio-visual efficient tuning methods.
Here, we additionally provide a comparison of \textit{inference}-time runtime.
The results are shown in Table~\ref{tab:inference_runtime_comparison}.
Compared to prior methods, our model consistently achieves significantly lower inference runtime across three audiovisual learning tasks.

\bibliography{egbib}
\bibliographystyle{iclr2022_conference}

\end{document}

%% file: macros.tex
\usepackage{amsthm}
\usepackage[english]{babel}
\usepackage[T1]{fontenc}
\usepackage{amsmath}
\usepackage{amssymb}
\usepackage{amsfonts}
\usepackage{multirow}
\usepackage{mathtools}
\usepackage{breqn}
\usepackage{bbm}
\usepackage{dsfont}
\usepackage{graphicx}
\usepackage{natbib}
\usepackage{cases}
\usepackage[colorinlistoftodos]{todonotes}

\newtheorem*{remark*}{Remark}

\newtheorem*{observation*}{Observation}

\numberwithin{equation}{section}

\newcommand{\Gnorm}[1]{{\left\vert\kern-0.25ex\left\vert\kern-0.25ex\left\vert #1 
		\right\vert\kern-0.25ex\right\vert\kern-0.25ex\right\vert}}
\newcommand{\gnorm}[1]{{\vert\kern-0.25ex\vert\kern-0.25ex\vert #1 
		\vert\kern-0.25ex\vert\kern-0.25ex\vert}}

%% file: tables/sota_ave.tex
\begin{table*}[t]
    \belowrulesep=0pt
    \aboverulesep=0pt
    \caption{
    \textbf{Comparison on audio-visual event localization task.} We evaluate our model using both modality-shared and modality-separate transformer backbones. In various setups, our method has competitive accuracy while significantly reducing memory and runtime costs. 
    }
    \centering
    \resizebox{1.0\textwidth}{!}{
    \begin{tabular}{l|cccc|cc|cc|c}
        \hline
     {Method} & {\shortstack{Visual \\ Encoder} }
       & \shortstack{Audio \\ Encoder} 
       & \shortstack{Visual \\ Pretrain Dataset} 
       & \shortstack{Audio \\ Pretrain Dataset} 
       & \shortstack{Trainable \\ Params (\%) $\downarrow$} 
       & \shortstack{Total \\ Params (M) $\downarrow$} 
       & \shortstack{Memory \\ (GB) $\downarrow$} 
       & \shortstack{Runtime \\ (ms) $\downarrow$} 
       & Acc $\uparrow$ \\
       \hline 
   
       AVEL~\cite{tian2018audio}                & ResNet-152                & VGGish           & ImageNet     & AudioSet            & 2.7     & 136.0     & N/A & N/A  & 74.0 \\
       AVSDN~\cite{lin2019dual}                 & ResNet-152               & VGGish          & ImageNet     & AudioSet            & 5.7  &    140.3  & N/A & N/A& 75.4 \\
       CMRAN~\cite{xu2020cross}         & ResNet-152                & VGGish           & ImageNet     & AudioSet            & 10.7     & 148.2  & N/A & N/A    & 78.3 \\
       MM-Pyramid~\cite{yu2021mm}         & ResNet-152                & VGGish           & ImageNet     & AudioSet            & 25.0     & 176.3 & N/A & N/A    & 77.8 \\
       CMBS~\cite{xia2022cross}              & ResNet-152                & VGGish           & ImageNet     & AudioSet  & 6.6 & 216.7 &N/A & N/A  & 79.7 \\
       \midrule
       LAVisH~\cite{lin2023lavish}          & \multicolumn{2}{c}{ViT-L-16 ~(shared)}   & ImageNet & \ding{56}           & 4.3   & 340.1   & 19.2 & 328.3 & 78.1 \\
       CoPL~\cite{zhao2024CoPL}          & \multicolumn{2}{c}{ViT-L-16 ~(shared)}   & ImageNet & \ding{56}           & 1.5 &  332.8 &  N/A & N/A & 79.2 \\
       AVMoE~\cite{cheng2024avmoe}          & \multicolumn{2}{c}{ViT-L-16 ~(shared)}   & ImageNet & \ding{56}           & 32.6 & 483.1   & 19.5 & 461.5 & 79.2 \\
       MA-AVT~\cite{mahmud2024maavt}          & \multicolumn{2}{c}{ViT-L-16 ~(shared)}   & ImageNet & \ding{56}           & 3.7 & 338.4   & 7.2 & 448.5 & 79.6 \\
       
       \textit{\textbf{Mettle (ours)}}  & \multicolumn{2}{c}{ViT-L-16 ~(shared)}   & ImageNet & \ding{56}           & 38.8 & 532.2   & 1.7 & 146.2 & 80.6 \\
       \midrule
       %
      LAVisH~\cite{lin2023lavish}         & \multicolumn{2}{c}{Swin-V2-L ~(shared) }             & ImageNet & \ding{56}          & 2.7   & 238.8 & 15.4 & 322.1 &  81.1 \\
      AVMoE~\cite{cheng2024avmoe}         & \multicolumn{2}{c}{Swin-V2-L ~(shared) }             & ImageNet & \ding{56}   & 39.4 & 374.4  & 16.8 & 500.0 & 81.5 \\
      STG-CMA~\cite{wang2024stgcma}         & \multicolumn{2}{c}{Swin-V2-L ~(shared) }             & ImageNet & \ding{56}   & 8.9 & 214.0  & 5.9 & 429.5 & 82.5 \\
      \textit{\textbf{Mettle (ours)}}        & \multicolumn{2}{c}{Swin-V2-L ~(shared) }             & ImageNet & \ding{56}          & 37.2   & 364.4   & 1.8 & 125.4 & 82.3 \\
      \midrule
      LAVisH~\cite{lin2023lavish} & {Swin-V2-L}  & HTS-AT  & ImageNet & AudioSet         & 30.6  & 374.9 & 11.9 & 232.9 &  78.6 \\
      DG-SCT~\cite{duan2024dgsct} & {Swin-V2-L}  & HTS-AT  & ImageNet & AudioSet         & 43.6  & 461.3 & 13.1 & 316.7 &  82.2 \\
      AVMoE~\cite{cheng2024avmoe} & {Swin-V2-L}  & HTS-AT  & ImageNet & AudioSet         & 34.9  & 404.0 & 12.1 & 256.1 &  82.6 \\
      \textit{\textbf{Mettle (ours)}} & {Swin-V2-L}  & HTS-AT  & ImageNet & AudioSet         & \textbf{23.0}  & \textbf{338.0 }& \textbf{1.5} & \textbf{88.3} &  \textbf{83.3} \\
      
       \bottomrule
    \end{tabular}
    }
   \label{tab:sota_ave}
\end{table*}

%% file: tables/sota_avvp.tex
\begin{table*}
    \belowrulesep=0pt
    \aboverulesep=0pt
    \caption{\textbf{Comparison on the audio-visual video parsing task.} `Avg.' is the average value of segment-level and event-level metrics. Swin-V2-L~\cite{liu2022swinv2} and HTS-AT~\cite{chen2022htsat} are used as the visual and audio transformer backbones, respectively.
    }
    \centering
    \resizebox{\linewidth}{!}{
    \begin{tabular}{l|cc|cc|ccccc| ccccc|c}
        \toprule
        \multirow{2}{*}{Method} 
& \multirow{2}{*}{\shortstack{Trainable \\ Params (\%) $\downarrow$}} 
& \multirow{2}{*}{\shortstack{Total \\ Params (M) $\downarrow$}} 
& \multirow{2}{*}{\shortstack{Memory \\ (GB) $\downarrow$}} 
& \multirow{2}{*}{\shortstack{Runtime \\ (ms) $\downarrow$}} 
& \multicolumn{5}{c|}{Segment-level $\uparrow$} 
& \multicolumn{5}{c|}{Event-level $\uparrow$} 
& \multirow{2}{*}{Avg. $\uparrow$} \\

        &&&& & A & V & AV & Type & Event & A & V & AV & Type & Event & \\
        \midrule
        HAN~\cite{tian2020unified} & N/A & N/A& N/A & N/A & 60.1 & 52.9 & 48.9 & 54.0 & 55.4 & 51.3 & 48.9 & 43.0 & 47.7 & 48.0 & 51.0 \\
        MGN~\cite{mo2022multi}& N/A & N/A& N/A & N/A & 60.7 & 55.5 & 50.6 & 55.6 & 57.2 & 51.0 & 52.4 & 44.4 & 49.3 & 49.2 & 52.6 \\ \midrule
        DG-SCT~\cite{duan2024dgsct} & 43.3 & 458.9 & 8.5 & 299.3 & 59.0 & 59.4 & 52.8 & 57.1 & 57.0 & 49.2 & 56.1 & 46.1 & 50.5 & 49.1 & 53.6  \\
        AVMoE~\cite{cheng2024avmoe} & 30.9 & 376.4 & 12.9 & 490.0 & 62.1 & \textbf{60.0} & \textbf{54.4} & \textbf{58.8} & 59.0 & 51.8 & \textbf{55.7} & \textbf{47.6} & \textbf{51.7} & 50.2 & \textbf{55.1} \\
        \textit{\textbf{Mettle (ours)}} & \textbf{8.3} & \textbf{283.5} & \textbf{0.6} & \textbf{64.2} & \textbf{64.3} & 55.9 & 50.9 & 57.0 & \textbf{61.2} & \textbf{56.1 }& 52.2 & 45.1 & 51.1 & \textbf{53.1} & 54.7 \\
        \bottomrule
    \end{tabular}
    }
    \label{tab:sota_avvp}
\end{table*}

%% file: tables/sota_avs.tex
\begin{table*}[t]
    \belowrulesep=0pt
    \aboverulesep=0pt
    \caption{
    \textbf{Comparison on the audio-visual segmentation task.} We evaluate our approach on both the single-source (S4) and multi-source (MS3) segmentation problems. 
    }
    \label{tab:sota_avs}
    \centering
    \resizebox{1\textwidth}{!}{
    \begin{tabular}{l| cccc | cc| cc| cc}
        \toprule
Method & \shortstack{Visual \\ Encoder} 
       & \shortstack{Audio \\ Encoder} 
       & \shortstack{Visual \\ Pretrain Dataset} 
       & \shortstack{Audio \\ Pretrain Dataset} 
       & \shortstack{Trainable \\ Params (\%) $\downarrow$} 
       & \shortstack{Total \\ Params (M) $\downarrow$} 
       & \shortstack{Memory \\ (GB) $\downarrow$} 
       & \shortstack{Runtime \\ (ms) $\downarrow$} 
       & \shortstack{S4 \\ (mIoU) $\uparrow$} 
       & \shortstack{MS3 \\ (mIoU) $\uparrow$} \\

        \midrule
        AVS~\cite{zhou2022avs}   & PVT-V2  & VGGish   & ImageNet&AudioSet & 58.7 & 174.5  & 4.8 & 647.0 & 78.7 & 54.0\\
        \midrule
         LAVisH~\cite{lin2023lavish}  & \multicolumn{2}{c}{Swin-V2-L ~(shared) }  & ImageNet & \ding{56}&  14.0 & 266.4 & 5.9 & 222.6 & 80.1 & 49.8 \\
         \textit{\textbf{Mettle (ours)}}   & \multicolumn{2}{c}{Swin-V2-L ~(shared)}  & ImageNet & \ding{56}&  11.1 & 257.2 & 2.7 & 113.2 & 79.9 & 53.7 \\
         \midrule
         DG-SCT~\cite{duan2024dgsct}  & {Swin-V2-L } & HTS-AT & ImageNet & AudioSet  & 61.5 & 594.8 & 9.0 & 330.3 & 80.9 & 53.5 \\
         AVMoE~\cite{cheng2024avmoe}  & {Swin-V2-L } & HTS-AT & ImageNet & AudioSet  & 54.4 & 501.2 & 9.5 & 427.3 & \textbf{81.1} & 54.5 \\
         \textit{\textbf{Mettle (ours)}}  & {Swin-V2-L } & HTS-AT & ImageNet & AudioSet  & \textbf{21.6 }& \textbf{291.7} &\textbf{2.9}& \textbf{105.7} & 80.7 & \textbf{55.1} \\
         
        \bottomrule
    \end{tabular}
}
\end{table*}

%% file: tables/tab_compare_with_other_memory_efficient_methods.tex
\begin{table*}[t]
    \belowrulesep=0pt
    \aboverulesep=0pt
    
\caption{\textbf{Comparison with state-of-the-art memory-efficient methods from vision-language field.} Our method consistently outperforms them in
accuracy on both tasks, demonstrating a better trade-off between accuracy and efficiency.}
\centering
\tiny
\resizebox{1\linewidth}{!}{
\begin{tabular}{c|c|ccc|c|ccc|c}
\hline
\multirow{2}{*}{Method} & \multirow{2}{*}{Venue} & \multicolumn{4}{c|}{AVEL} & \multicolumn{4}{c}{AVVP}  \\ \cmidrule(r){3-10}
& & \shortstack{Total \\ Params (M) $\downarrow$} 
& \shortstack{Memory \\ (GB) $\downarrow$} 
& \shortstack{Runtime \\ (ms) $\downarrow$} 
& Acc. $\uparrow$ 
& \shortstack{Total \\ Params (M) $\downarrow$} 
& \shortstack{Memory \\ (GB) $\downarrow$} 
& \shortstack{Runtime \\ (ms) $\downarrow$} 
& Avg. $\uparrow$ \\

    \midrule
    {LST}~\cite{sung2022lst} & NeurIPS'22 & 347.8 & 1.3 & {79.0} & 80.0   & 286.1 & {0.6} & \textbf{59.0} & 51.9 \\
    {UniPT}~\cite{diao2024unipt} & CVPR'24 &  343.2 & 1.3 & \textbf{83.9} & 81.1    & 285.5 &0.6&61.2&52.4 \\
    SHERL~\cite{diao2024sherl} & ECCV'24 &  347.7 & \textbf{1.0} & 90.7 & 81.6   &286.0&{0.6}&71.7&52.1  \\ \midrule
    \textbf{\textit{Mettle (ours)}} & - & \textbf{338.0 }& 1.5 & 88.3 & \textbf{83.3}   & \textbf{283.5} & \textbf{0.6} & 64.2 & \textbf{54.7} \\
        \hline
\end{tabular}
}
\label{tab:comparison_with_memory_efficient_methods}
\end{table*}

%% file: tables/supp_tab_compare_with_frozen_linearprob.tex
\begin{table}[t]
    \belowrulesep=0pt
    \aboverulesep=0pt
    \centering
    \caption{\textbf{Comparison with two vanilla approaches.} `Direct' refers to using features extracted by frozen transformer models directly for downstream tasks. `Linear' represents the linear probing strategy, where a modality-separate linear layer is added after the final frozen transformer layer.}
    \label{tab:comparison_with_linearprobing} 
    \tiny
    \resizebox{0.7\linewidth}{!}{  
        \begin{tabular}{c|c|cc|cc|c|cc}
            \toprule
            \multirow{3}{*}{Method} & AVEL & \multicolumn{5}{c|}{AVVP} & \multicolumn{2}{c}{AVS} \\ 
            \cmidrule(r){2-9}
             & \multirow{2}{*}{Acc.} & \multicolumn{2}{c|}{Segment-level} & \multicolumn{2}{c|}{Event-level} & \multirow{2}{*}{Avg.} & {S4} & MS3 \\
            & & Type & Event & Type & Event & & mIoU & mIoU \\
            \midrule
            Direct & 80.7 &  55.3&57.0&49.0&49.1&52.3     & 78.8&48.7 \\
            Linear & 78.8 & 54.8&57.1&48.3&49.5&51.9     &79.2&51.8\\ \midrule
            \textbf{Ours} & \textbf{82.3}   & \textbf{57.0}&\textbf{61.2}&\textbf{51.1}&\textbf{53.1}&\textbf{54.7}   & \textbf{79.9}&\textbf{53.7}\\
            \bottomrule
        \end{tabular}
    }
\end{table}

%% file: tables/Ab_meta_token_and_step_num.tex
\begin{table}[t]
    \belowrulesep=0pt
    \aboverulesep=0pt
    \centering
    \caption{\textbf{Effect of the number of meta tokens $K$.} We analyze how varying the number of audio ($K_a$) and visual ($K_v$) meta tokens affects performance.}
    \footnotesize
    \resizebox{0.7\linewidth}{!}{
    \begin{tabular}{cc| c |cc|cc|c |cc }
        \toprule
        \multirow{3}{*}{$K_a$} & \multirow{3}{*}{$K_v$} & AVEL & \multicolumn{5}{c|}{AVVP} & \multicolumn{2}{c}{AVS} \\ \cmidrule(r){3-10}
        & & \multirow{2}{*}{Acc.} & \multicolumn{2}{c|}{Segment-level} & \multicolumn{2}{c|}{Event-level} & \multirow{2}{*}{Avg.} & S4 & MS3 \\
        & & & Type & Event & Type & Event & & mIoU & mIoU \\
        \midrule
        1 & 1 & \textbf{82.3}   & 57.0 & 61.2 & 51.1 & 53.1 & \textbf{54.7} & \textbf{79.9} & \textbf{53.7} \\ 
        1 & 2 & 81.2 & 57.0 & 62.0 & 50.6 & 52.8 & 54.5 & 79.6 & 52.7 \\
        2 & 1 & 81.2 & 56.6 & 60.7 & 50.4 & 52.3 & 54.1 & 79.4 & 52.0 \\
        2 & 2 & 80.7 & 56.5 & 60.6 & 50.3 & 52.1 & 54.0 & 79.7 & 52.6 \\
        4 & 4 & 80.8 & 56.1 & 60.4 & 49.7 & 51.9 & 53.6 & 79.8 & 52.3 \\
        \bottomrule
    \end{tabular}
    }
    \label{tab:meta_token_number}
\end{table}

%% file: tables/supp_hierarical_metatoken_numers.tex
\begin{table}[t]\footnotesize
    \belowrulesep=0pt
    \aboverulesep=0pt
    \centering
    \caption{\textbf{Effect of using hierarchical meta-token numbers.} Swin-V2-L~\cite{liu2022swinv2} and HTS-AT~\cite{chen2022htsat}, both featuring hierarchical designs, are used as the visual and audio backbones, respectively.  Each backbone consists of four-stage transformer layers. The first column presents different choices for the meta-token numbers at four stages (\eg, the \{8,4,2,1\} indicates that the layers in the first bottom stage use 8 meta-tokens while layers of the fourth top stage use 1 meta-token). The `default' setting denotes that all transformer stages/layers share the same meta-token numbers.}
    \label{tab:hierarchical_metatoken_numers} 
    \tiny
    \resizebox{0.7\linewidth}{!}{  
        \begin{tabular}{c|c|cc|cc|c}
            \toprule
            \multirow{3}{*}{Setup} & AVEL & \multicolumn{5}{c}{AVVP} \\ 
            \cmidrule(r){2-7}
             & \multirow{2}{*}{Acc.} & \multicolumn{2}{c|}{Segment-level} & \multicolumn{2}{c|}{Event-level} & \multirow{2}{*}{Avg.} \\
            & & Type & Event & Type & Event & \\
            \midrule
            \{8,4,2,1\} & 80.7 &  55.3&57.0&49.0&49.1&52.3   \\
            \{2,2,1,1\} & 78.8 & 54.8&57.1&48.3&49.5&51.9   \\ \midrule
            default & \textbf{82.3}   & \textbf{57.0}&\textbf{61.2}&\textbf{51.1}&\textbf{53.1}&\textbf{54.7} \\
            \bottomrule
        \end{tabular}
    }
\end{table}

%% file: tables/Ab_distill_step_number.tex
\begin{table}[t]
    \belowrulesep=0pt
    \aboverulesep=0pt
    \centering
    \caption{\textbf{Effect of the number of distillation steps $R$.} We study the impact of varying the number of audio ($R_a$) and visual ($R_v$) distillation iterations.}
    \resizebox{0.7\linewidth}{!}{
    \begin{tabular}{cc| c |cc|cc|c |cc }
        \toprule
        \multirow{3}{*}{$R_a$} & \multirow{3}{*}{$R_v$} & AVEL & \multicolumn{5}{c|}{AVVP} & \multicolumn{2}{c}{AVS} \\ \cmidrule(r){3-10}
        & & \multirow{2}{*}{Acc.} & \multicolumn{2}{c|}{Segment-level} & \multicolumn{2}{c|}{Event-level} & \multirow{2}{*}{Avg.} & S4 & MS3 \\
        & & & Type & Event & Type & Event & & mIoU & mIoU \\
        \midrule
        1 & 1 & \textbf{82.3}   & 57.0 & 61.2 & 51.1 & 53.1 & \textbf{54.7} & 79.6 & \textbf{53.7} \\ 
        1 & 2 & 80.9 & 55.8 & 60.9 & 49.5 & 52.0 & 53.4 & \textbf{79.9} & 53.5 \\
        2 & 1 & 80.7 & 56.7 & 61.2 & 50.4 & 52.5 & 54.2 & \textbf{79.9} & \textbf{53.7} \\
        2 & 2 & 81.1 & 56.4 & 60.6 & 50.0 & 51.9 & 53.8 & 79.8 & 53.2 \\
        4 & 4 & 81.6 & 56.6 & 61.1 & 50.3 & 52.6 & 54.1 & 79.8 & 52.3 \\
        \bottomrule
    \end{tabular}
    }
    \label{tab:distill_step_number}
\end{table}

%% file: tables/Ab_component_of_lcd_mti.tex
\begin{table}[t]
    \belowrulesep=0pt
    \aboverulesep=0pt
    \centering
    \caption{\textbf{Component ablation of LCD.} `TS' represents the Task-Specific meta-token tuning branch, and `DP' denotes the linear branch for preserving Domain-Pretrained knowledge.}
    \resizebox{0.7\linewidth}{!}{
    \begin{tabular}{cc| c |cc|cc|c |cc }
        \toprule
        \multicolumn{2}{c|}{Component}  & AVEL & \multicolumn{5}{c|}{AVVP} & \multicolumn{2}{c}{AVS} \\ \cmidrule(r){1-10}
        \multirow{2}{*}{TS} &\multirow{2}{*}{DP} & \multirow{2}{*}{Acc.} & \multicolumn{2}{c|}{Segment-level} & \multicolumn{2}{c|}{Event-level} & \multirow{2}{*}{Avg.} & S4 & MS3 \\
         & & & Type & Event & Type & Event & & mIoU & mIoU \\
        \midrule
        \ding{56} & \ding{52} & 77.3 & 55.3 & 60.1 & 49.0 & 51.8 & 52.9 & 78.9 & 52.7 \\ 
        \ding{52} & \ding{56} & 80.5 & 56.2 & 60.8 & 49.7 & 51.6 & 53.6 & 78.6 & 51.5 \\ 
        \ding{52} & \ding{52} & \textbf{82.3} & \textbf{57.0} & \textbf{61.2} & \textbf{51.1} & \textbf{53.1} & \textbf{54.7} & \textbf{79.9} & \textbf{53.7} \\ 
        \bottomrule
    \end{tabular}
    }
    \label{tab:component_for_LCD}
\end{table}

%% file: tables/supp_using_avgpooling_branch_in_LCD.tex
\begin{table}[t]
    \belowrulesep=0pt
    \aboverulesep=0pt
    
    \centering
    \caption{\textbf{Replacing the linear layer branch of the LCD module with simple average pooling (`AvgPool.').} The item `Remove' denotes that this branch is completely removed. Swin-V2-L~\cite{liu2022swinv2} and HTS-AT~\cite{chen2022htsat} are used as the visual and audio transformer backbones, respectively.}
    \label{tab:using_avgpooling_in_LCD}
    \resizebox{0.7\linewidth}{!}{  
        \begin{tabular}{c|c|cc|cc|c|cc}
            \toprule
            \multirow{3}{*}{Strategy} & AVEL & \multicolumn{5}{c|}{AVVP} & \multicolumn{2}{c}{AVS} \\ 
            \cmidrule(r){2-9}
             & \multirow{2}{*}{Acc.} & \multicolumn{2}{c|}{Segment-level} & \multicolumn{2}{c|}{Event-level} & \multirow{2}{*}{Avg.} & {S4} & MS3 \\
            & & Type & Event & Type & Event & & mIoU & mIoU \\
            \midrule
            Remove & {80.5}   & 56.2&60.8&49.7&51.6&{53.6}   & 78.6&{51.5}\\ 
            AvgPool. & 83.2 &  56.4&59.8&49.4&50.4&53.3     &80.4&53.8 \\ \midrule
            \textbf{Linear} & \textbf{83.3}   & \textbf{57.0}&\textbf{61.2}&\textbf{51.1}&\textbf{53.1}&\textbf{54.7}   & \textbf{80.7}&\textbf{55.1}\\
            \bottomrule
        \end{tabular}
    }
\end{table}

%% file: tables/Ab_component_of_mti.tex
\begin{table}[t]
    \belowrulesep=0pt
    \aboverulesep=0pt
    \centering
    \caption{\textbf{Component ablation of MTI.} MTI module involves the intra- and cross-modality interactions.}
    \begin{tabular}{cc|cc}
        \toprule
        \multicolumn{2}{c|}{Component} & \multicolumn{2}{c}{AVS} \\
        intra & cross & S4 & MS3 \\
        \midrule
        \ding{52} & \ding{56} & 79.3 & 49.9 \\
        \ding{56} & \ding{52} & 79.4 & 52.3 \\
        \ding{52} & \ding{52} & \textbf{79.9} & \textbf{53.7} \\
        \bottomrule
    \end{tabular}
    \label{tab:component_for_MTI}
\end{table}

%% file: tables/tab_training_configures.tex
\begin{table}[hp]
    \belowrulesep=0pt
    \aboverulesep=0pt
    \caption{\textbf{Training hyperparameters for audiovisual downstream tasks}
    }
    \centering
    \begin{tabular}{l|c|c|c|c}
        \toprule
         Hyperparameters & AVEL & AVVP & AVS-S4 & AVS-MS3  \\
        \midrule
        batch size & 16 & 16 & 4 & 4 \\
        epochs & 10 & 15 & 40 & 30 \\
        optimizer & Adam & Adam & Adam & Adam \\
        base learning rate & 4e-4 & 3e-4 & 1e-4 & 1e-4 \\
        learning rate scheduler & StepLR & StepLR & StepLR & None \\
        \multicolumn{1}{r|}{$\hookrightarrow$ \textnormal{step size}} & 3 & 5 & 5 & N/A \\
        \multicolumn{1}{r|}{$\hookrightarrow$ \textnormal{gamma}} & 0.35 & 0.5 & 0.8 & N/A \\
        \bottomrule
    \end{tabular}
    \label{tab:supp_training_hyperparameters}
\end{table}

%% file: tables/supp_hyperparameter_for_modality_specific_backbones.tex
\begin{table}[hp]
    \belowrulesep=0pt
    \aboverulesep=0pt
    \centering
    \caption{\textbf{Hyperparameter details} when using modality-specific (Swin-V2-L~\cite{liu2022swinv2} for visual, HTS-AT~\cite{chen2022htsat} for audio) transformer backbones for model training.}
    \label{tab:supp_parameters_for_modality_specific_backbones} 
        \begin{tabular}{c|cc|cc}
            \toprule
            \multirow{2}{*}{Tasks} & \multicolumn{2}{c|}{Meta-token number $K$} & \multicolumn{2}{c}{Distillation step $R$} \\ \cmidrule(r){2-5}
            & $K_a$ & $K_v$ & $R_a$ & $R_v$ \\ \midrule
            AVEL & 1 & 2 & 1 & 1 \\
            AVVP & 1 & 1 & 1 & 1 \\
            AVS (S4) & 1 & 1 & 1 & 2 \\
            AVS (MS3) & 1 & 1 & 1 & 1 \\
            \bottomrule
        \end{tabular}
\end{table}